\documentclass[letterpaper, 10 pt, conference]{ieeeconf}  
\usepackage{amsmath}
\usepackage{booktabs}
\usepackage{multirow}
\usepackage{xcolor}
\usepackage{colortbl}
\usepackage{tabularray}
\usepackage{cite}
\usepackage{placeins}
\usepackage{booktabs,threeparttable}
\usepackage{graphicx} 
\bstctlcite{BSTcontrol}
\usepackage{etoolbox}
\usepackage{capt-of} 

\definecolor{darkgreen}{RGB}{0, 154, 85}
\definecolor{orange}{RGB}{255, 128, 0}

\usepackage[colorlinks=true,
            linkcolor=orange,
            citecolor=orange,
            urlcolor=orange]{hyperref}
\usepackage{cleveref}
\crefname{figure}{\textcolor{orange}{Fig.}}{\textcolor{orange}{Figs.}}
\Crefname{figure}{\textcolor{orange}{Fig.}}{\textcolor{orange}{Figs.}}

\crefname{table}{\textcolor{orange}{Table}}{\textcolor{orange}{Tables}}
\Crefname{table}{\textcolor{orange}{Table}}{\textcolor{orange}{Tables}}

\crefname{section}{\textcolor{orange}{Section}}{\textcolor{orange}{Sections}}
\Crefname{section}{\textcolor{orange}{Section}}{\textcolor{orange}{Sections}}

\crefname{equation}{\textcolor{orange}{Eq.}}{\textcolor{orange}{Eqs.}}
\Crefname{equation}{\textcolor{orange}{Eq.}}{\textcolor{orange}{Eqs.}}
            
\IEEEoverridecommandlockouts                              

\usepackage{etoolbox} 

\newcommand{\myfigure}{%
    \begin{center}
        \vspace{0.6em}
        \includegraphics[width=0.985\linewidth]{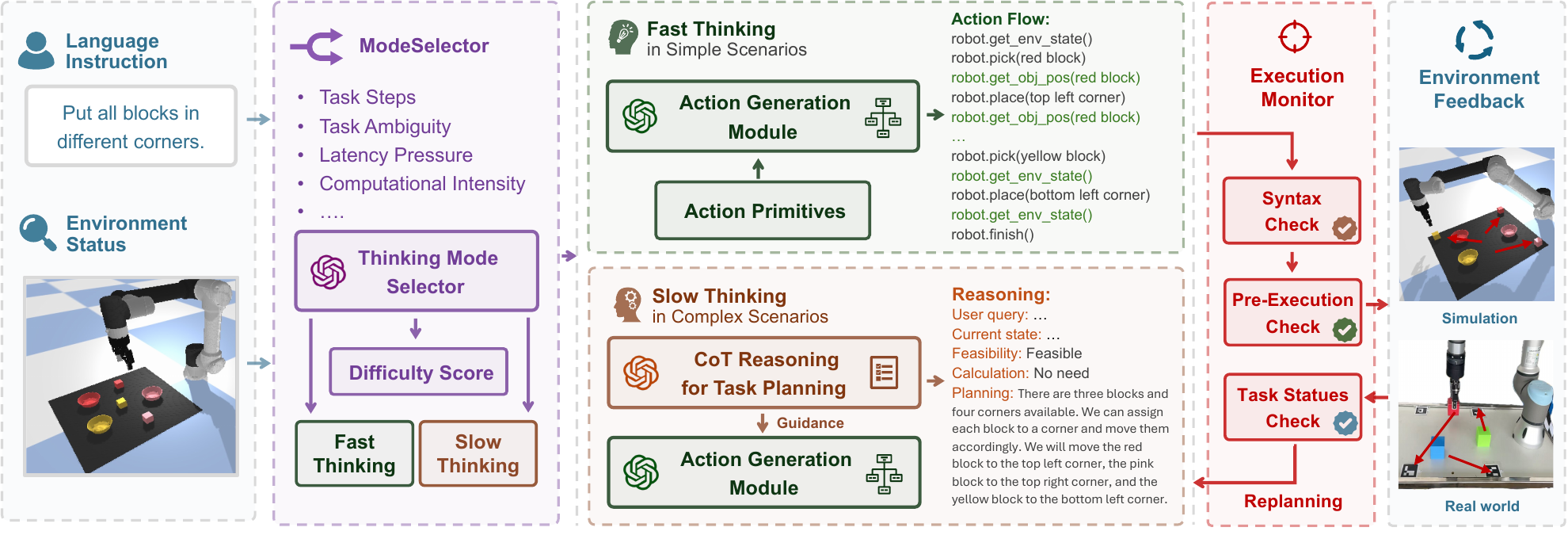}
        \vspace{-0.6em}
        \captionof{figure}{RoboPilot is a dual-thinking closed-loop agentic system for dynamic manipulation, enabling replanning with adaptive dual-thinking for dynamic environments, while balancing the manipulation efficiency and accuracy. Mode selector processes language instruction and visual information to select fast- or slow-thinking mode. The action generation module then orchestrates action primitives to solve the task, guided by the Chain-of-Thought reasoning module (only in slow-thinking mode). The execution monitor validates generated actions and tracks environment changes, providing feedback for replanning.
}
        \label{fig:robopilot_framework}
        \vspace{-0.7em} 
    \end{center}
}

\makeatletter
\patchcmd{\@maketitle} 
  {\end{center}}      
  {\end{center}\myfigure} 
  {}{}                 
\makeatother

\title{\LARGE \bf
RoboPilot: Generalizable Dynamic Robotic Manipulation \\ with Dual-thinking System\vspace{-0.7em}}

\author{Xinyi Liu$^{1,2}$, Mohammadreza Fani Sani$^{1}$, Julius Wirbel$^{2}$, Bahram Zarrin$^{1}$, Zewei Zhou$^{3\dagger}$ and Roberto Galeazzi$^{2}$
\\ [0.1cm]
\small \tt{\href{https://sherryliu3670.github.io/robopilot/}{https://sherryliu3670.github.io/robopilot/}}
\thanks{$^{1}$Xinyi Liu, Mohammadreza Fani Sani, and Bahram Zarrin are with Microsoft. {\tt\small liusherry@microsoft.com}}%
\thanks{$^{2}$Xinyi Liu, Julius Wirbel, and Roberto Galeazzi are with the Technical University of Denmark. $^{3}$Zewei Zhou is with University of California, Los Angeles {\tt\small zeweizhou@ucla.edu}. $^{\dagger}$Corresponding author.} 
}

\begin{document}

\maketitle
\thispagestyle{empty}
\pagestyle{empty}
\begin{abstract}

Despite rapid progress in robotics, complex or long-horizon tasks remain a fundamental challenge. Most current approaches follow an open-loop paradigm with limited reasoning and no feedback, resulting in poor robustness to environmental changes and severe error accumulation. We present \textit{RoboPilot}, a dual-thinking closed-loop agentic framework for robotic manipulation that supports adaptive reasoning for complex tasks in real-world dynamic environments. RoboPilot leverages primitive actions for structured task planning and flexible action generation as a agentic system, while introducing feedback to enable replanning from dynamic changes and execution errors. Chain-of-Thought reasoning further enhances high-level task planning and guides low-level action generation. The agentic system dynamically switches between fast and slow thinking to balance efficiency and accuracy. To systematically evaluate the robustness of RoboPilot in diverse robot manipulation scenarios, we introduce \textit{RoboPilot-Bench}, a benchmark spanning 21 tasks across 10 categories, including infeasible-task recognition and dynamic recovery. Experiments show that RoboPilot outperforms state-of-the-art baselines by 11\% in task success rate, and the real-world deployment on an industrial robot further demonstrates its robustness.
\end{abstract}

\section{Introduction}
General-purpose robots, also known as generalist robots, have emerged as a central focus in robotics research due to their potential to autonomously execute diverse tasks in unseen real-world environments \cite{kim2024review}. Recent advances in large language models (LLMs) and agentic systems have demonstrated remarkable capabilities in mapping natural language instructions to robot planning and control \cite{lv2024robomp, mann2020language}. By transferring the extensive world knowledge and reasoning capability of the LLM models to the physical world, the integration of LLMs into robotic systems offers a powerful framework to streamline the planning and execution of robotic tasks. Despite this, real-world complex or long-horizon tasks remain challenging for robot manipulation \cite{zhang2025generative, liu2025dynamem}, as they require robust execution under dynamic conditions, and powerful and adaptive reasoning capability. 

These requirements manifest as two fundamental challenges: \textbf{\textit{1) Static planning without closed feedback
loop for dynamic changes}}. Previous work focused on static planning, where a single plan is generated at the beginning of the task without subsequent replanning \cite{liang2022code, arenas2024prompt, jin2024robotgpt}. This paradigm often leads to task failure due to accumulated errors or suboptimal plans, particularly in complex or long-horizon scenarios, and it cannot adapt to unexpected situations, such as execution failures or changes in object positions. \textbf{\textit{2) Lack of strong and adaptive reasoning capabilities to handle various complex and long-horizon tasks}}. Robots need to decompose problems into stepwise solutions and solve them with strong reasoning capability, yet early studies relied on single-pass action generation within LLMs \cite{huang2023instruct2act, zhu2024language, rana2023sayplan}, which limits adaptability and impedes replanning in dynamic environments.

In this work, we introduce \textbf{RoboPilot}, a dual-thinking closed-loop agentic robo-system that solves real-world manipulation tasks in dynamic environments. In contrast to prior approaches that rely on elaborate prompt engineering with manipulation examples \cite{liang2022code, arenas2024prompt, gupta2023visual}, our system uses action primitives as abstracted API functions with formal structures as the tools in the agentic system, breaking down complex manipulation tasks into high-level task planning and low-level action parameter generation. 
Rather than statically generating actions \cite{jin2024robotgpt, singh2022progprompt, qiu2024open}, RoboPilot continuously monitors task progress, integrates environment feedback, and leverages historical messages to recover from dynamic changes and execution errors. Furthermore, to enhance reasoning over complex or long-horizon tasks, we incorporate Chain-of-Thought (CoT) reasoning to support complex computation and explicitly decouple high-level task planning from low-level action generation as separate steps. To avoid unnecessary reasoning in simple scenarios, we introduce an LLM-based thinking mode selector that chooses between a fast-thinking mode and a CoT-enhanced slow-thinking mode based on the task complexity.

\textbf{\textit{The lack of benchmarks that evaluate robotic systems in dynamic and long-horizon tasks}} remains a critical challenge. Existing manipulation benchmarks prioritize object diversity \cite{luo2025fmb} or task generalization \cite{zakka2025mujoco}, but fail to test for manipulation robustness in dynamic situations. To address this gap, we introduce a manipulation benchmark consisting of 10 groups for a total of 21 tasks, including long-horizon planning, infeasible objectives, and deliberately designed failure cases. Our experimental results, both in simulation and with the real-world robot, demonstrate that RoboPilot achieves substantial improvements over state-of-the-art (SOTA) baselines, exhibiting strong robustness in dynamic environments. In summary, our main contributions are:

\begin{itemize}
    \item We propose RoboPilot, a dual-thinking closed-loop agentic system for dynamic manipulation, enabling replanning with adaptive dual-thinking modes for dynamic environments, particularly in complex tasks.  
    \item We adopt primitive actions to structure the task planning and action generation as the tools, and introduce the feedback and replanning module to facilitate recovery from dynamic changes and errors. Chain-of-Thought reasoning is introduced in slow-thinking to enhance task planning and guide low-level action generation. RoboPilot can further dynamically switch between dual-thinking modes, balancing efficiency and accuracy.
    \item We present \textit{RoboPilot-Bench}, a comprehensive benchmark for robotics manipulation, covering infeasible task recognition and dynamic recovery for robustness test.  
    \item RoboPilot achieves an overall 11\% improvement in success rate over state-of-the-art methods, and real-world deployment further demonstrates its robustness.
\end{itemize}

\section{Related Work}

\noindent \textbf{Large Language Models for Dynamic Robotics Manipulation.}
LLM‐based robotic manipulation aims to couple language understanding, task planning, and action generation to generalize to unseen scenarios, while traditional robotic control focuses on feedback and closed-loop replanning \cite{zhang2025generative, de2012theory}. However, many recent LLM-based approaches choose to adopt a static, one-shot paradigm for planning and action generation, overlooking the role of feedback \cite{liang2022code, arenas2024prompt, jin2024robotgpt,gupta2023visual}. Several works have explored LLM-based closed-loop manipulation, but focus on high-level task planning recovery with limited attention to whether generated actions can reliably accomplish the intended tasks. Therefore, tight integration of task planning with action-level replanning and enhancing the reasoning capabilities of LLMs with feedback to handle complex replanning scenarios remains underexplored \cite{liu2023reflect, yao2023react,han2024interpret}. Moreover, step-by-step reasoning with Chain-of-Thought offers a promising avenue to tackle these challenges. However, most CoT-based manipulation frameworks remain largely confined to static planning \cite{zhu2024language, zawalski2024robotic, zhang2024learning}, leaving CoT reasoning for dynamic manipulation and robustness enhancement underexplored as well.

\noindent \textbf{Robotics Manipulation Benchmarks.} 
Comprehensive benchmarks are fundamental for evaluating and comparing model performance \cite{wang2025large}. Existing robotic manipulation benchmarks have made significant progress in task and object diversity \cite{james2020rlbench, jiang2022vima, zakka2025mujoco, brohan2022rt}, and some have further aimed to assess generalization and knowledge transfer \cite{liu2023libero}. While few benchmarks address long-horizon manipulation \cite{mees2022calvin}, a comprehensive evaluation of robustness under real-world, non-stationary conditions, spanning replanning and dynamic recovery, remains an open challenge.

\section{Methodology}

This section introduces the \textit{RoboPilot} framework for closed-loop dynamic manipulation with dual-thinking modes, as illustrated in \cref{fig:robopilot_framework}. First, \cref{sec:fast_thinking} explains the fast thinking mode with single-stage generation for actions and the closed-loop replanning mechanism. Second, \cref{sec:slow_thinking} presents the slow-thinking mode with CoT reasoning for task planning to guide the action generation. Last, \cref{sec:dual_thinking} introduces the dual-thinking system and \cref{sec:implementation_details} provides the implementation details.

\setcounter{figure}{1}
\begin{figure}[t]
    \centering
    \vspace{0.2cm}
    \includegraphics[width=0.98\linewidth]{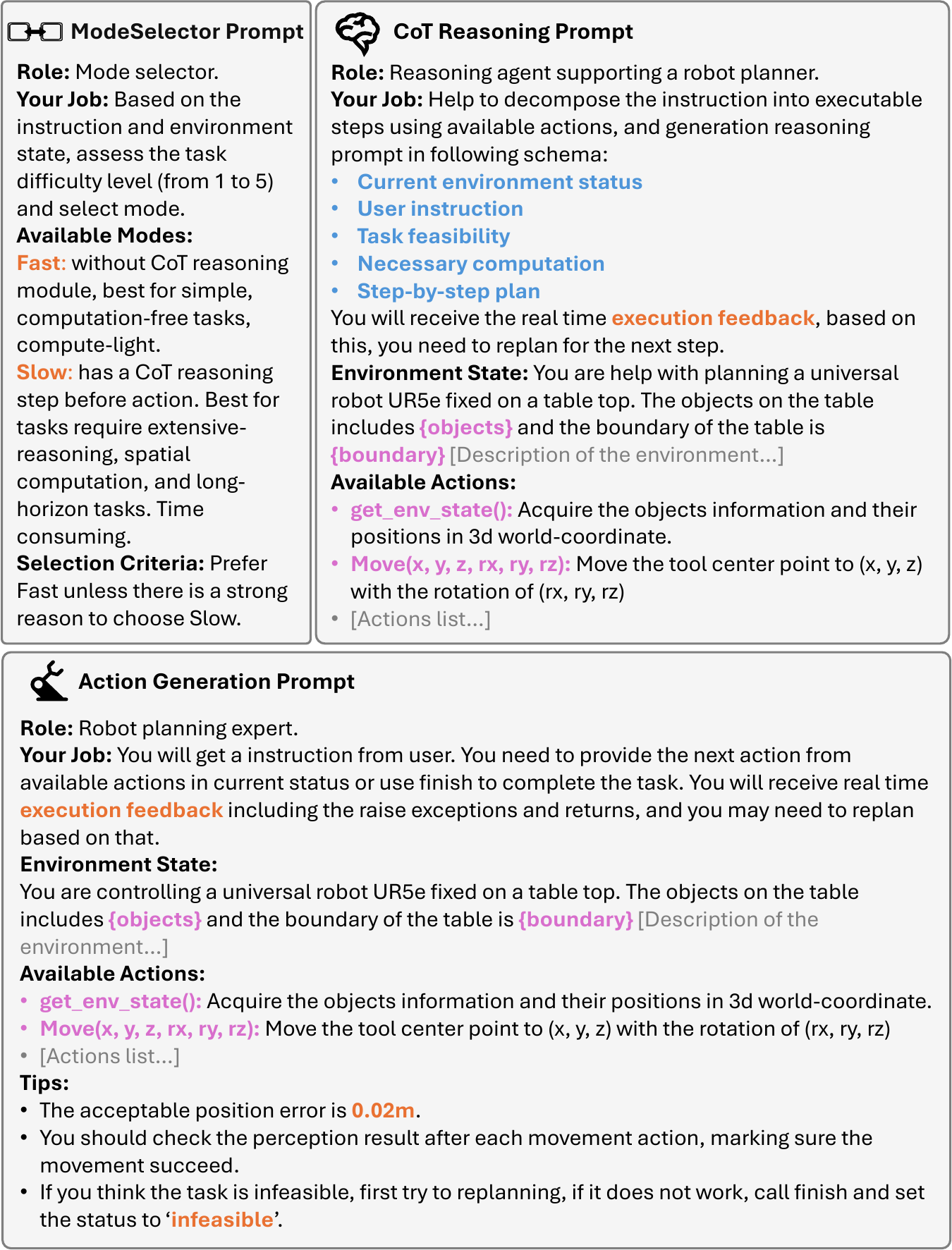}
    \caption{Prompt Snapshot for RoboPilot. The prompt includes three parts for the ModeSelector, CoT Reasoning, and Action Generation module. The prompt used in the fast-thinking mode is identical to that of the slow-thinking mode, but omits the reasoning prompt.}
    \vspace{-0.5cm}
    \label{fig:prompt}
\end{figure}

\subsection{Fast-Thinking RoboPilot (RoboPilot FT)}
\label{sec:fast_thinking}

RoboPilot FT integrates task planning and action generation through structured action primitives in single-stage generation, while introducing a closed-loop replanning mechanism for dynamic manipulation, achieving fast and robust performance in simple dynamic tasks.

\noindent \textbf{Structured Reasoning with Action Primitives.} RoboPilot FT processes instructions in natural language and vision information for task planning and action generation, which first coordinates high-level task decomposition and action orchestration, then generates low-level actions. Prior work has primarily focused on generating code \cite{liang2022code, arenas2024prompt, huang2023instruct2act}, often resulting in the generation of similar actions (e.g., \texttt{pick} and \texttt{place}). This ultimately demands extensive human-crafted prompts for task planning and action generation to ensure generation stability. In contrast, RoboPilot defines fundamental manipulation skills as action primitives with corresponding functions. This design structures task planning as a composition of action primitives, and action generation as parameter instantiation of these primitives, thereby unifying task planning and action generation within a single reasoning framework. 

\noindent \textbf{Closed-loop Replanning for Dynamic Manipulation.} After each task planning and action generation loop, the Execution Monitor conducts pre-execution validity checks to capture infeasible action calls and parameter assignments to prevent unsafe executions. Following every action, the monitor integrates environment feedback into the history messages as closed-loop feedback. It explicitly evaluates execution status, especially for movement primitives, by comparing the object position after execution to the intended target. If the deviation exceeds a predefined threshold, recovery is triggered through a system message. Based on this updated execution feedback, history messages, and the new environment state, the system performs replanning to generate subsequent plans for the remaining part of the task. The loop continues until the model explicitly issues a \texttt{finish} operator. 

The system maintains an explicit, interpretable plan–action memory, so replanning reduces to extending or locally editing the same structured trace rather than regenerating monolithic and brittle code.

\subsection{Slow-Thinking RoboPilot (RoboPilot ST)}
\label{sec:slow_thinking}

While RoboPilot-FT is efficient, its single-stage planning–action generation can fail on complex or long-horizon problems, e.g., scenarios that are spatially computation-intensive or require conditional reasoning. To strengthen reasoning capability, we introduce a slow-thinking mode that enhances task planning with CoT reasoning, decoupling task planning from action generation into two explicit stages. 

\noindent \textbf{Task Planning with CoT Reasoning.} 
We integrate CoT reasoning into task planning to strengthen high-level, long-horizon task decomposition. A step-by-step CoT mechanism equips the agent with the reasoning capability required for complex and long-horizon tasks, while also tackling dynamic environments that demand inter-step history tracing and on-the-fly replanning. The CoT planner invokes a \texttt{get\_reasoning} operator conditioned on the user’s language instruction and visual context, producing a stepwise rationale that produces (i) current environment status; (ii) user instruction (iii) task feasibility (iv) related calculation (v) a step-by-step plan for action primitives orchestration.
\noindent \textbf{Action Generation.} The CoT reasoning results are feed into the action-generation module, ensuring that subsequent operator selections are consistent with the intermediate analysis. The action generation module adopts the same action primitives as in the fast thinking mode.

\noindent \textbf{Closed-loop Replanning for Dynamic Manipulation.} By decoupling planning and action, the slow-thinking mode enables high-level task replanning and low-level recovery in action generation. 
After each execution, feedback from environments is injected into the reasoning stream as assistant messages, enabling cross-module grounding.

\subsection{Dual-Thinking Mechanism}
\label{sec:dual_thinking}

To trade off the performance and time-efficiency, we introduce a prompt-based ModeSelector, using LLM to zero-shot analyze the instructions and environment states to select the RoboPilot FT or ST mode. As shown in \cref{fig:prompt}, based on a series of parameters including number of task steps, requirement for spacial reasoning, task ambiguity and required timeframe for a solution. The ModeSelector is also prompted with key aspects to consider in mode selection as well as information for conflict resolution.

\FloatBarrier
\begin{figure*}[t]
    \centering
    \vspace{0.2cm}
    \includegraphics[width=0.96\linewidth]{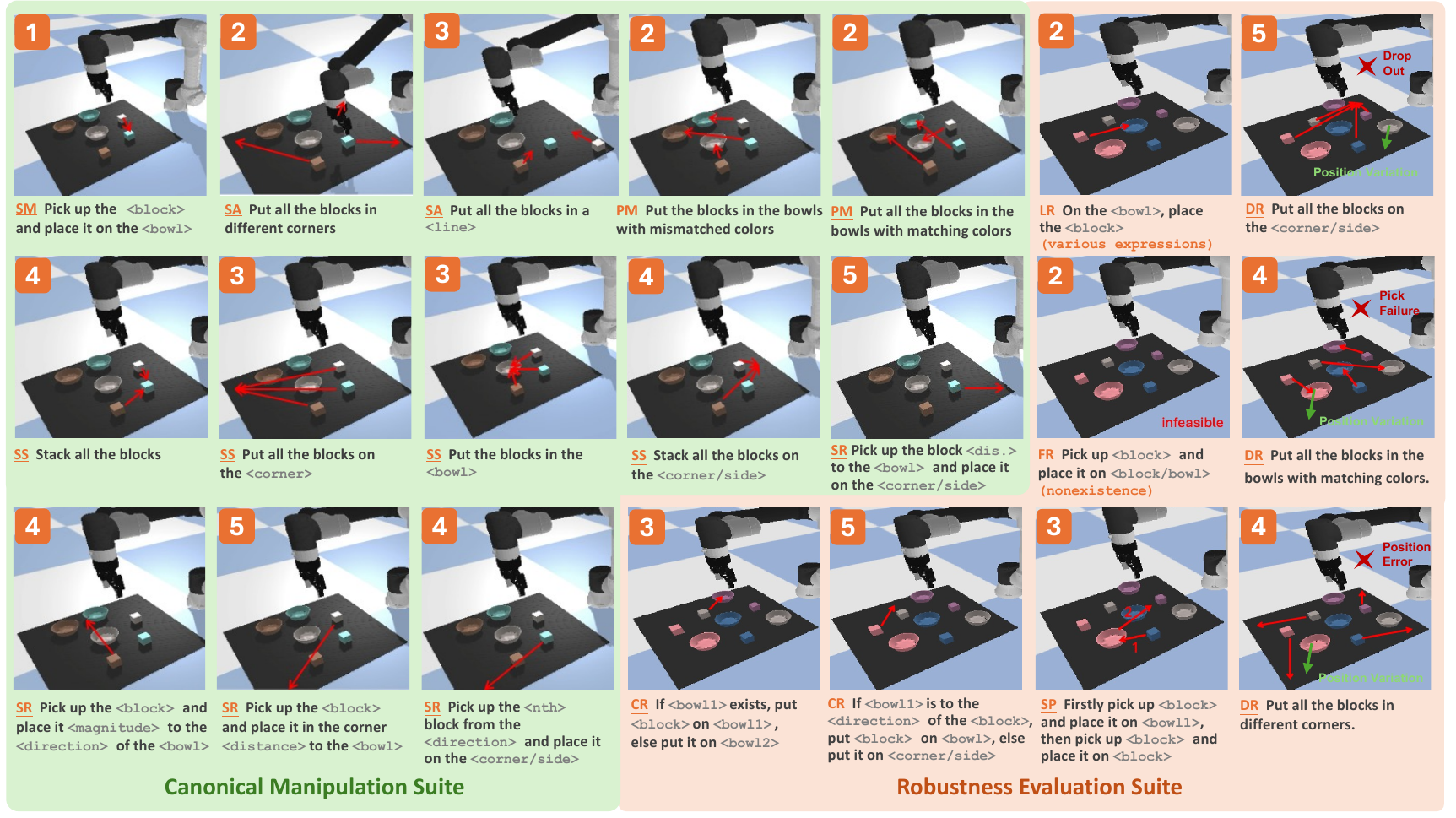}
    \caption{Task definition and difficulty in RoboPilot-Bench, including the Canonical Manipulation Suite (5 Groups with 13 Tasks) and Robustness Evaluation Suite (5 Groups with 8 Tasks). A three-pair example is provided with red arrows highlighting one possible solution. The difficulty score between 1 (easy) and 5 (hard) is shown in the top left corner.}
    \vspace{-0.5cm}
    \label{fig:benchmark}
\end{figure*}

\subsection{Implementation Details}
\label{sec:implementation_details}

\noindent \textbf{Prompt:} We designed a structured prompting strategy for each module in RoboPilot, shown in \cref{fig:prompt}. Each prompt consists of the role definition, job description, and specific information and requirements for each module. 

\noindent \textbf{Action Primitives:} Action primitives are abstracted as function APIs including all necessary low-level skills for dynamic manipulation, and divided into two categories: \textit{1) Perception Primitives}, invoking onboard cameras and acquire visual information of the environment, utilized during every replanning step; and \textit{2) Execution Primitives}, corresponding to fundamental robot actions for object grasping and movement.

\section{Benchmark}

To evaluate the performance of dynamic manipulation with a focus on robustness, we propose RoboPilot-Bench, illustrated in \cref{fig:benchmark}. This benchmark is composed of \textit{(i) the Canonical Manipulation Suite}, and \textit{(ii) the Robustness Evaluation Suite}, which is new, extended task suite with categories targeting robustness evaluation such as infeasible task recognition and dynamic recovery tasks.

\subsection{Canonical Manipulation Suite}
The Canonical Manipulation Suite follows the benchmark setting of \cite{liang2022code} and provides a set of 13 tabletop manipulation tasks involving blocks and bowls with five categories:

\noindent \textbf{\textit{1) Simple Manipulation (SM)}}: Basic pick-and-place actions testing low-level action execution.

\noindent \textbf{\noindent \textit{2) Spatial Allocation (SA)}}: Long-horizon tasks requiring spatial allocation and constraints.

\noindent \textbf{\noindent \textit{3) Stable Stacking (SS)}}: Long-horizon tasks requiring vertical stacking blocks, sensitive to physical stability.

\noindent \textbf{\noindent \textit{4) Perceptual Matching (PM)}}: Tasks requiring perceptual grounding by associating blocks with bowls via appearance.

\noindent \textbf{\noindent \textit{5) Spatial Reasoning (SR)}}: Reasoning-challenging tasks demanding explicit spatial computing and relational inference (e.g., relative distances and ordered selection).

\begin{table*}[t]
\centering
\renewcommand{\arraystretch}{1.15}
\setlength{\tabcolsep}{1.9pt}
\begin{threeparttable}
\vspace{0.2cm}
\caption{Average Success Rate (\%) over Task Groups in RoboPilot-Bench  (50 trials per task)}
\label{fig:performance}
\begin{tabular}{l|c|c|c|c|c|c|c|c|c|c|c}
\toprule
Method       & \multicolumn{1}{l|}{\begin{tabular}[c]{@{}l@{}}Simple \\Manipulation\end{tabular}} 
             & \multicolumn{1}{l|}{\begin{tabular}[c]{@{}l@{}}Spatial \\Allocation\end{tabular}} 
             & \multicolumn{1}{l|}{\begin{tabular}[c]{@{}l@{}}Stable \\Stacking\end{tabular}} 
             & \begin{tabular}[c]{@{}l@{}}Perceptural \\Matching\end{tabular} 
             & \begin{tabular}[c]{@{}l@{}}Spatial \\Reasoning\end{tabular} 
             & \begin{tabular}[c]{@{}l@{}}Conditional \\Reasoning\end{tabular} 
             & \begin{tabular}[c]{@{}l@{}}Sequential \\Planning\end{tabular} 
             & \begin{tabular}[c]{@{}l@{}}Linguistic \\Robustness\end{tabular} 
             & \begin{tabular}[c]{@{}l@{}}Avg\end{tabular} 
             & \begin{tabular}[c]{@{}l@{}}Feasibility \\Recognition\end{tabular}  
             & \begin{tabular}[c]{@{}l@{}}Dynamic \\Recovery \end{tabular} \\ 
\midrule
CaP$^{*}$ \cite{liang2022code}         
&  94 &  93 & 59  & 99  & 52  &  16 & 86  & 100  & 74.9  & -  &  - \\
PromptBook$^{*}$ \cite{arenas2024prompt}   
&  98 &  93 & 67  & 98  &  60 &  32 & 92  &  100 & 80.0  &  - & -  \\
Instruct2Act$^{*}$ \cite{huang2023instruct2act} 
& 96  & 94  &  71 &  98 & 62  & 62  &  95 &  100 & 84.8  &  - & - \\ 
\midrule
RoboPilot FT 
& 98  &  91 & 87  & 98  & 79  &  77 & 100  & 100  &  91.3 & 94  & 87 \\
RoboPilot ST 
& 98  & 97  & 88  &  100 & 91  & 95  & 100  &  100 &  96.1 & 94  & 86  \\
RoboPilot (w/o replan)
& 96  & 92  & 67  &  100 & 85  & 92  & 100  &  100 &  91.5 & 94  & 0 \\
\rowcolor{gray!20} \textbf{RoboPilot}    
&   \textbf{98} \tiny{\textcolor{orange}{+0}}  
&   \textbf{97} \tiny{\textcolor{orange}{+3}}  
&   \textbf{87} \tiny{\textcolor{orange}{+16}}  
&   \textbf{100} \tiny{\textcolor{orange}{+1}}  
&   \textbf{89} \tiny{\textcolor{orange}{+27}}  
&   \textbf{95} \tiny{\textcolor{orange}{+33}}  
&   \textbf{100} \tiny{\textcolor{orange}{+5}}  
&   \textbf{100} \tiny{\textcolor{orange}{+0}}  
&   \textbf{95.8} \tiny{\textcolor{orange}{+11}}  
&   \textbf{94}   
&   \textbf{86}  \\
\bottomrule
\end{tabular}
\end{threeparttable}
\vspace{-0.2cm}
\end{table*}

\vspace{0.1cm}

\begin{table}[t]
\centering
\renewcommand{\arraystretch}{1.15}
\setlength{\tabcolsep}{5.5pt}
\caption{Real-world Experiment Success Rate (\%) (10 trials per group).}
\label{tab:realworld}
\begin{tabular}{l|c|c|c|c|c|c|c|c|c}
\toprule
Method & SA & PM & SR & CR & SP & FR & LR & DR & Avg. \\
\midrule
RoboPilot & 80 & 80 & 60 & 70 & 90 & 100 & 90 & 60 & 78.8 \\
\bottomrule
\end{tabular}
\end{table}

\begin{table}[t]
\centering
\renewcommand{\arraystretch}{1.1}
\setlength{\tabcolsep}{10pt}
\caption{RoboPilot Performance Comparison with Different Large Language Models}
\label{tab:foundation models}
\begin{tabular}{l|c|c} 
\toprule
Method & Success Rate (\%) & Avg. Time per Step (s)  \\ 
\midrule
Deepseek-R1 \cite{guo2025deepseek}       &      93.2             &           9.20              \\
GPT-5 \cite{openai_gpt5_system_card_2025}       &           \textbf{95.8}        &            11.86             \\
\rowcolor{gray!20} GPT-4o \cite{hurst2024gpt}       &         92.5          &            \textbf{5.61}             \\
\bottomrule
\end{tabular}
\vspace{-0.3cm}
\end{table}

\begin{table}[t]
\centering
\renewcommand{\arraystretch}{1.15}
\setlength{\tabcolsep}{7.7pt}
\begin{threeparttable}
\caption{Average Time Cost and Tokens in RoboPilot-Bench}
\label{tab:time cost}
\begin{tabular}{l|c|c} 
\toprule
Method & Avg. Time per Step (s) &  Avg. Input Tokens  \\ 
\midrule
CaP$^{*}$\cite{liang2022code}          &  5.63 &   3850  \\
PromptBook$^{*}$ \cite{arenas2024prompt} &  6.21 &    4680  \\
Instruct2Act$^{*}$ \cite{huang2023instruct2act}   &  7.11 & 2720  \\ 
\midrule
RoboPilot FT & 4.06  &   1400   \\
RoboPilot ST &  7.63 &   1740   \\
\rowcolor{gray!20} \textbf{RoboPilot} &  \textbf{5.61} \tiny{\textcolor{orange}{-0.02}} \tiny{(ModeSelector cost 0.63)} &   \textbf{1970} \tiny{\textcolor{orange}{-1880}}   \\
\bottomrule
\end{tabular}
\end{threeparttable}
\vspace{-0.3cm}
\end{table}

\subsection{Robustness Evaluation Suite}
We further introduce the Robustness Evaluation Suite, which focus on robustness testing for dynamic tasks. It covers five additional task groups with 8 tasks:  

\noindent \textbf{\textit{1) Conditional Reasoning (CR)}}: Context-dependent tasks requiring conditional reasoning over spatial relations, posing significant reasoning challenges.

\noindent \textbf{\textit{2) Sequential Planning (SP)}}: Multi-step tasks with strict temporal dependencies, explicitly evaluating long-horizon and order-sensitive task.

\noindent \textbf{\textit{3) Feasibility Recognition (FR)}}: Unsolvable tasks designed to assess the agent’s ability to detect infeasibility and avoid redundant actions. The task feasibility label will be provided as the evaluation reference.

\noindent \textbf{\textit{4) Linguistic Robustness (LR)}}: Language-varied queries assessing robustness to syntactic and semantic paraphrasing in natural language instructions.

\noindent \textbf{\textit{5) Dynamic Recovery (DR)}}: Long-horizon tasks reflecting real-world dynamics, where stochastic execution failures and object pose variations necessitate adaptive recovery and replanning strategies.

\subsection{Difficulty} 
We provide a relative difficulty rating for the proposed benchmark. Following the grading methods of \cite{pan2024dynathink, jin2024robotgpt}, we base the rating on the performance of state-of-the-art methods such as Code-as-Policies\cite{liang2022code}, PromptBook\cite{arenas2024prompt}, and Instruct2Act\cite{huang2023instruct2act}. The final difficulty scores are further calibrated by considering the number of steps and the computational and reasoning complexity involved. 

The benchmark difficulty score is defined as the base score of a task under three block–bowl pairs. Since the evaluation suite varies the number of block–bowl pairs between two and four, task difficulty is scaled accordingly. We define the score at three objects as the representative difficulty for each task. This score is given to all tasks, regardless the number of objects involved.

\section{Experiments}
The section provides details on experiment settings and three sets of experiment results: First, we compare RoboPilot against state-of-the-art baselines in both performance and robustness. Second, we evaluate its efficiency in simulation, measuring both time and computational cost. Lastly, we assess RoboPilot performance on a real-world robot platform.

\subsection{Experiment Settings}
To evaluate the proposed framework under as realistic conditions as possible, ten randomized scenarios are created for the benchmark runs. Each scenario contains between two and four block-bowl pairs with uniformly sampled colors and initial poses. For each scenario, each task from both suites are executed five times with independently sampled object and spacial goals, resulting in 50 distinct trials per task.

\noindent \textbf{LLM Models.} 
We standardize the LLM backbone to GPT-4o~\cite{hurst2024gpt}, with a temperature setting of $0$ to ensure reproducible output. As RoboPilot operates in an interactive closed loop, a hard budget cap of 20 LLM invocations per trial is imposed to prevent query loops. Furthermore, performance comparisons with GPT-5 and Deepseek-R1 are performed.

\begin{figure*}
    \centering
    \vspace{0.2cm}
    \includegraphics[width=1\linewidth]{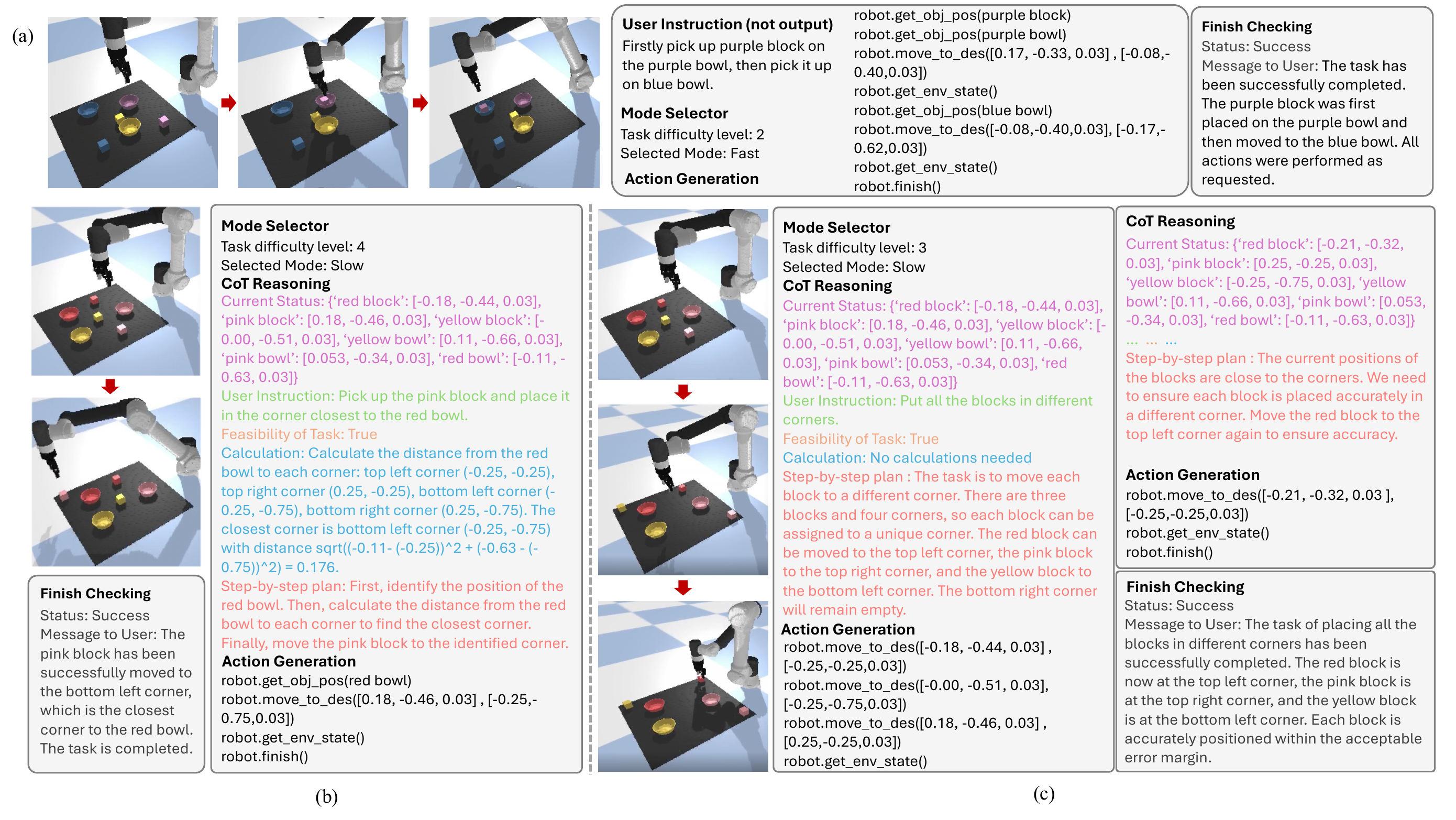}
    \vspace{-0.5cm}
    \caption{Qualitative Experimental Results: (a) Simple sequential reasoning task (Fast-Thinking Mode). (b) A spatial reasoning task (Slow-Thinking Mode). (c) An dynamic recovery long-horizon task (Slow-Thinking Mode).}
    \label{fig:output}
    \vspace{-0.5cm}
\end{figure*}

    \noindent \textbf{Baseline Comparisons. } To form the baseline for our performance comparisons, we reimplemented  Code as Policies (CaP) \cite{liang2022code}, PromptBook \cite{arenas2024prompt}, and Instruct2Act \cite{huang2023instruct2act}, integrating GPT-4o as their foundation model. These reimplementations are shown as CaP$^{*}$, PromptBook$^{*}$ and Instruct2Act$^{*}$.

\noindent \textbf{Metrics.} 
To evaluate the performance, a success rate based on the final position of the objects is used. Given the goal status $S_{\text{goal}}\in\mathcal{G}=\{S_{\text{goal}_1},\dots,S_{\text{goal}_m}\}$, task success is defined by \cref{eq:success} for all objects $o\in\mathcal{O}$.
\begin{equation}
\big\|S_{\text{actual}}(o)-S_{\text{goal}}(o)\big\|_2 \le \delta,
\label{eq:success}
\end{equation}
where the threshold distance $\delta=0.02\,\text{m}$ that adapt to the object size (block edge legth and bowl radius are both 0.05m). For feasibility recognition tasks, the agent’s final status prediction (\textit{success}/ \textit{failure}/ \textit{infeasible}) is compared to the ground-truth scenario feasibility label for success evaluation.

To evaluate the efficiency performance, we employ the Average Time per Step and Average Input Tokens representing inference time and throughput respectively. The step here is defined as successfully moving one object.

\noindent \textbf{Robot Platforms.} In simulation, we use a UR5e robotic arm in a tabletop manipulation environment based on PyBullet. For the real-world experiment, we adpot a UR3e robotic arm. The Robotiq 2F85 gripper is employed in all experiments.

\subsection{Main Results}
\noindent \textbf{Simulation Results.} \cref{fig:performance} reports average success over the ten task groups included in the benchmark. RoboPilot achieves an overall success rate of \textbf{95.8\%}, surpassing the strongest baseline Instruct2Act by \textbf{11\%}, while also enable task feasibility identification and replanning in dynamic environment. 
The results in the benchmark highlight four main observations:
\paragraph{Strong Robustness with Closed-loop Replanning}
        Unexpected changes and execution error always lead to manipulation failure for the static planning manipulation system, results in all baselines without replanning struggling with long-horizon tasks (\emph{Stable Stacking}) and are entirely incapable of addressing \emph{Dynamic Recovery} scenarios. In contrast, RoboPilot's closed-loop re-planing framework improves the \emph{Stable Stacking} success rate to 87\%, also achieves 86\% success rate in dynamic tasks requiring re-planing. This performance gap is further elucidated by the ablation results, where disabling the replanning module leads to a 20\% decline in \emph{Stable Stacking} and a total system failure in \emph{Dynamic Recovery} (86\% to 0\%), demonstrating replanning is indispensable for robustness against execution failures.
\paragraph{Action Primitives Benefit Reasoning-Intensive Tasks}
RoboPilot FT improves the average success rate by 11\% over the strongest baseline, with the largest gains observed in \emph{Spatial Reasoning} and \emph{Stable Stacking} tasks. This improvement arises because code-based approaches are brittle, an error in the generated code can cause the entire system to fail. Moreover, the absence of structured reasoning for task planning and action generation makes such methods less effective on reasoning-intensive tasks.

\paragraph{CoT Reasoning for Complex Tasks}
    For tasks demanding multi-step inference and complex computation, RoboPilot ST with CoT reasoning significantly outperforms both RoboPilot FT and all other baselines, reaching 91\% success rate on \emph{Spatial Reasoning} tasks. Furthermore, RoboPilot ST achieves 95\% success rate on \emph{Conditional Reasoning} tasks, comparing to 77\% for RoboPilot FT. These results support the hypothesis that explicit CoT reasoning improves performance on spatial understanding and reasoning.

\begin{figure}[t]
    \centering
    \begin{minipage}{\columnwidth}
        \centering
        \includegraphics[width=0.77\linewidth]{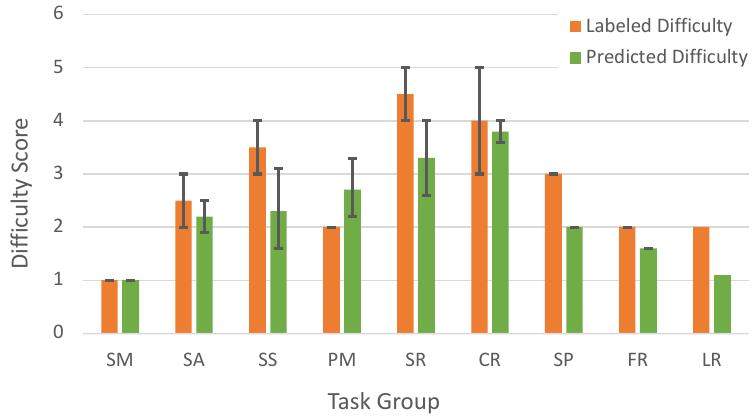}
        \vspace{-2pt}
        
        {\footnotesize (a) Comparison of Labeled and ModeSelector Predicted Task Difficulties on RoboPilot-Bench}
    \end{minipage}

    \vspace{4pt}
    \begin{minipage}{0.65\columnwidth}
        \centering
        \includegraphics[width=\linewidth]{picture/difficulty_mode.pdf}
        \vspace{-2pt}
        {\footnotesize (b) Relationship between Predicted Task Difficulty and Mode Selection}
    \end{minipage}
    \caption{ModeSelector Performance Analysis}
    \label{fig:mode select}
    \vspace{-0.7cm}
\end{figure}

\paragraph{Dual-thinking Efficiency–Accuracy Trade-off}
    The performance of RoboPilot with dual-thinking mode closely follows RoboPilot ST on reasoning-intensive tasks with same performance on Conditional Reasoning task and minimaly lower performance on Spacial Reasoning tasks, indicating the dual-thinking system reliably selects the slow-thinking mode when deeper reasoning is warranted. Later in \cref{fig:mode select}, we will further discuss that the selector prefers FT mode on simpler tasks, yielding comparable accuracy while substantially reducing inference time relative to ST in \cref{tab:time cost}.

\FloatBarrier
\begin{figure*}[!t]
    \centering
    \vspace{0.2cm}
    \includegraphics[width=0.97\linewidth]{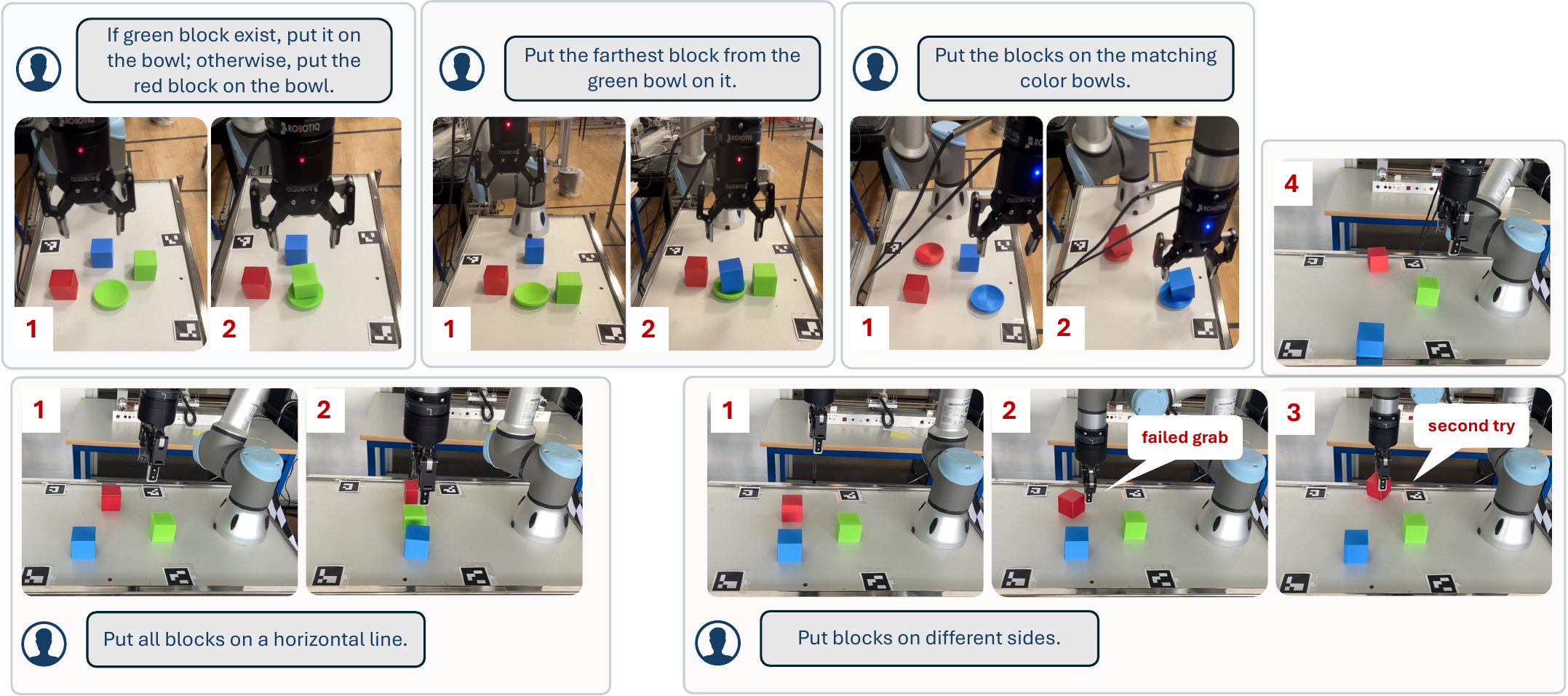}
    \caption{Real-world Experiment Results on UR3E Robot. Our RoboPilot demonstrates its performance and robustness across diverse manipulation tasks, especially the dynamic recovery tasks.}
    \label{fig:ur3e}
    \vspace{-0.5cm}
\end{figure*}

\vspace{0.1cm}

\noindent \textbf{Real-World Results.} We evaluate RoboPilot on \textit{Robustness Evaluation Suite} and the challenging subsets (SA, PM, SR tasks) of the \textit{Canonical Manipulation Suite} (with the representative tasks for each group). Each task is executed 10 times.\cref{tab:realworld} reports the success rates.

RoboPilot achieves an average real-world success rate of \textbf{78.8\%}. This sim-to-real performance gap stems from unmodeled factors, including lighting variations, constrained workspaces, and real-world uncertainties. Despite these challenges, RoboPilot maintains strong performance on \emph{Sequential Planning} (90\%), \emph{Feasibility Recognition} (100\%), and \emph{Linguistic Robustness} (90\%), indicating that the system generalizes effectively to real-world setups. Conversely, \emph{Spatial Reasoning} and \emph{Conditional Reasoning} tasks decline of approximately 25\%, primarily attributed to perception deviations from camera depth measurements and calibration errors. Nevertheless, RoboPilot demonstrates robust real-world recovery from execution failures, achieving a 60\% success rate on \emph{Dynamic Recovery} due to its closed-loop replanning framework.

\noindent \textbf{ModeSelector Performance Evaluation.}
To assess the performance of the \textit{ModeSelector}, we analyze its predicted task difficulty and the selected mode over all RoboPilot-Bench groups. Each task is evaluated 10 times, and both task-level and group-level difficulties are averaged. \cref{fig:mode select} (a) compares the labeled difficulty with the predictions. These align closely with labels in relative ordering across groups based on the average difficulty scores and the error bars across tasks, demonstrating that the selector effectively captures comparative difficulty, even if absolute levels are conservatively estimated in most groups.

\cref{fig:mode select} (b) shows the probability that ModeSelector chooses slow-thinking versus the predicted task difficulty. The curve indicates that the expected decision boundary falls between 2.0 and 3.3, near the midpoint of the difficulty range. Reasoning-intensive tasks with average labeled difficulty scores of 4–4.5 are highly likely to be assigned to slow-thinking. 
This explains the similar performance between the dual-thinking and RoboPilot ST variants. Overall, the results indticate that the ModeSelector preserves relative difficulty and establishes a practical switching boundary, assigning slow-thinking mode to reasoning-intensive groups while retaining fast-thinking mode for simpler ones.

\noindent \textbf{LLMs Backbone Evaluation.}
We assess RoboPilot on RoboPilot-Bench with different LLMs backbones, shown in \cref{tab:foundation models} to compare accuracy and inference time. GPT-5 attains the highest success (95.8\%) but with the largest latency (11.86 s/step), while Deepseek-R1 reaches 93.2\% at a slightly faster speed. GPT-4o delivers 92.5\% at the fastest speed (5.61 s/step), offering the most favorable efficiency-accuracy trade-off.

\subsection{Efficiency Evaluation}
Efficiency is a crucial factor in robot manipulation, particularly in real-world settings, yet it is often overlooked in prior work.
\cref{tab:time cost} compares our framework to the selected baselines on inference time (avg. time per step) and throughput (avg. input tokens). Our framework achieves a lower average time per step, while using significantly fewer input tokens owing to the use of action primitives.

As expected, RoboPilot ST and Dual-Thinking require slightly more input tokens due to the additional prompts invoking the CoT reasoning. While RoboPilot ST shows notably higher average time per step compared to RoboPilot FT, the ModeSelector realizing a practical efficiency–accuracy trade-off by utilizing FT mode on simpler tasks.

\subsection{Qualitative Results}

We qualitatively analyze RoboPilot in simulation and real-world environments. \cref{fig:output} shows three simulated tasks: (a) sequential planning, where ModeSelector predicts low difficulty score and selects fast-thinking with closed-loop checking; (b) spatial reasoning, rated higher difficulty score, where slow-thinking with explicit computation guides action generation; (c) dynamic recovery, where a dropped block is detected, triggering CoT-guided replanning to ensure success.

\cref{fig:ur3e} shows some representative task examples in the real-world experiment, including conditional reasoning, spatial reasoning, perceptual matching, spatial allocation, and dynamic recovery tasks. RoboPilot shows strong robustness on these real-world manipulation tasks, and successfully recover from execution failures. Additional real-world experiment demos are provided in the supplementary video. 

\section{Conclusions}
In this paper, we present the \textit{RoboPilot}, a dual-thinking closed-loop agentic framework for robotic manipulation that supports adaptive reasoning for complex tasks in real-world dynamic environments. We adopt action primitives to structure the reasoning of task planning and action generation, and introduce Chain-of-Thought reasoning to enhance the task planning. We also introduce the ModeSelector to adaptively switch between the CoT-enhanced slow thinking mode and the non-CoT fast thinking mode. Furthermore, we propose a comprehensive manipulation benchmark, \textit{RoboPilot-Bench}, including infeasible objectives tasks, deliberately designed failure and dynamic recovery tasks, extending on prior benchmarks to test manipulation robustness. Our simulation and real-world experiments show a measurable improvement of our framework compared to other state-of-the-art approaches while also demonstrating the strong robustness in dynamic environments.




{
\renewcommand{\baselinestretch}{0.95}
\normalsize
\bibliographystyle{IEEEtran}
\bstctlcite{BSTcontrol}
\bibliography{reference}

\begin{thebibliography}{10}
\providecommand{\url}[1]{#1}
\csname url@samestyle\endcsname
\providecommand{\newblock}{\relax}
\providecommand{\bibinfo}[2]{#2}
\providecommand{\BIBentrySTDinterwordspacing}{\spaceskip=0pt\relax}
\providecommand{\BIBentryALTinterwordstretchfactor}{4}
\providecommand{\BIBentryALTinterwordspacing}{\spaceskip=\fontdimen2\font plus
\BIBentryALTinterwordstretchfactor\fontdimen3\font minus \fontdimen4\font\relax}
\providecommand{\BIBforeignlanguage}[2]{{%
\expandafter\ifx\csname l@#1\endcsname\relax
\typeout{** WARNING: IEEEtran.bst: No hyphenation pattern has been}%
\typeout{** loaded for the language `#1'. Using the pattern for}%
\typeout{** the default language instead.}%
\else
\language=\csname l@#1\endcsname
\fi
#2}}
\providecommand{\BIBdecl}{\relax}
\BIBdecl

\bibitem{kim2024review}
Y.~Kim, D.~Kim, J.~Choi, J.~Park, N.~Oh, and D.~Park, ``A survey on integration of large language models with intelligent robots,'' \emph{Intelligent Service Robotics}, vol.~17, pp. 1091--1107, 2024.

\bibitem{lv2024robomp}
Q.~Lv, H.~Li, X.~Deng, R.~Shao, M.~Y. Wang, and L.~Nie, ``Robomp $^2$: A robotic multimodal perception-planning framework with multimodal large language models,'' \emph{arXiv preprint arXiv:2404.04929}, 2024.

\bibitem{mann2020language}
B.~Mann, N.~Ryder, M.~Subbiah, J.~Kaplan, P.~Dhariwal, A.~Neelakantan, P.~Shyam, G.~Sastry, A.~Askell, S.~Agarwal \emph{et~al.}, ``Language models are few-shot learners,'' \emph{arXiv preprint arXiv:2005.14165}, vol.~1, no.~3, p.~3, 2020.

\bibitem{zhang2025generative}
K.~Zhang, P.~Yun, J.~Cen, J.~Cai, D.~Zhu, H.~Yuan, C.~Zhao, T.~Feng, M.~Y. Wang, Q.~Chen \emph{et~al.}, ``Generative artificial intelligence in robotic manipulation: A survey,'' \emph{arXiv preprint arXiv:2503.03464}, 2025.

\bibitem{liu2025dynamem}
P.~Liu, Z.~Guo, M.~Warke, S.~Chintala, C.~Paxton, N.~M.~M. Shafiullah, and L.~Pinto, ``Dynamem: Online dynamic spatio-semantic memory for open world mobile manipulation,'' in \emph{2025 IEEE International Conference on Robotics and Automation (ICRA)}.\hskip 1em plus 0.5em minus 0.4em\relax IEEE, 2025, pp. 13\,346--13\,355.

\bibitem{liang2022code}
J.~Liang, W.~Huang, F.~Xia, P.~Xu, K.~Hausman, B.~Ichter, P.~Florence, and A.~Zeng, ``Code as policies: Language model programs for embodied control,'' \emph{arXiv preprint arXiv:2209.07753}, 2022.

\bibitem{arenas2024prompt}
M.~G. Arenas, T.~Xiao, S.~Singh, V.~Jain, A.~Ren, Q.~Vuong, J.~Varley, A.~Herzog, I.~Leal, S.~Kirmani \emph{et~al.}, ``How to prompt your robot: A promptbook for manipulation skills with code as policies,'' in \emph{2024 IEEE International Conference on Robotics and Automation (ICRA)}.\hskip 1em plus 0.5em minus 0.4em\relax IEEE, 2024, pp. 4340--4348.

\bibitem{jin2024robotgpt}
Y.~Jin, D.~Li, J.~Shi, P.~Hao, F.~Sun, J.~Zhang, B.~Fang \emph{et~al.}, ``Robotgpt: Robot manipulation learning from chatgpt,'' \emph{IEEE Robotics and Automation Letters}, vol.~9, no.~3, pp. 2543--2550, 2024.

\bibitem{huang2023instruct2act}
S.~Huang, Z.~Jiang, H.~Dong, Y.~Qiao, P.~Gao, and H.~Li, ``Instruct2act: Mapping multi-modality instructions to robotic actions with large language model,'' \emph{arXiv preprint arXiv:2305.11176}, 2023.

\bibitem{zhu2024language}
M.~Zhu, Y.~Zhu, J.~Li, J.~Wen, Z.~Xu, Z.~Che, C.~Shen, Y.~Peng, D.~Liu, F.~Feng \emph{et~al.}, ``Language-conditioned robotic manipulation with fast and slow thinking,'' in \emph{2024 IEEE International Conference on Robotics and Automation (ICRA)}.\hskip 1em plus 0.5em minus 0.4em\relax IEEE, 2024, pp. 4333--4339.

\bibitem{rana2023sayplan}
K.~Rana, J.~Haviland, S.~Garg, J.~Abou-Chakra, I.~Reid, and N.~Suenderhauf, ``Sayplan: Grounding large language models using 3d scene graphs for scalable robot task planning,'' \emph{arXiv preprint arXiv:2307.06135}, 2023.

\bibitem{gupta2023visual}
T.~Gupta and A.~Kembhavi, ``Visual programming: Compositional visual reasoning without training,'' in \emph{Proceedings of the IEEE/CVF conference on computer vision and pattern recognition}, 2023, pp. 14\,953--14\,962.

\bibitem{singh2022progprompt}
I.~Singh, V.~Blukis, A.~Mousavian, A.~Goyal, D.~Xu, J.~Tremblay, D.~Fox, J.~Thomason, and A.~Garg, ``Progprompt: Generating situated robot task plans using large language models,'' \emph{arXiv preprint arXiv:2209.11302}, 2022.

\bibitem{qiu2024open}
D.~Qiu, W.~Ma, Z.~Pan, H.~Xiong, and J.~Liang, ``Open-vocabulary mobile manipulation in unseen dynamic environments with 3d semantic maps,'' \emph{arXiv preprint arXiv:2406.18115}, 2024.

\bibitem{luo2025fmb}
J.~Luo, C.~Xu, F.~Liu, L.~Tan, Z.~Lin, J.~Wu, P.~Abbeel, and S.~Levine, ``Fmb: a functional manipulation benchmark for generalizable robotic learning,'' \emph{The International Journal of Robotics Research}, vol.~44, no.~4, pp. 592--606, 2025.

\bibitem{zakka2025mujoco}
K.~Zakka, B.~Tabanpour, Q.~Liao, M.~Haiderbhai, S.~Holt, J.~Y. Luo, A.~Allshire, E.~Frey, K.~Sreenath, L.~A. Kahrs \emph{et~al.}, ``Mujoco playground,'' \emph{arXiv preprint arXiv:2502.08844}, 2025.

\bibitem{de2012theory}
C.~C. de~Wit, B.~Siciliano, and G.~Bastin, \emph{Theory of robot control}.\hskip 1em plus 0.5em minus 0.4em\relax Springer Science \& Business Media, 2012.

\bibitem{liu2023reflect}
Z.~Liu, A.~Bahety, and S.~Song, ``Reflect: Summarizing robot experiences for failure explanation and correction,'' \emph{arXiv preprint arXiv:2306.15724}, 2023.

\bibitem{yao2023react}
S.~Yao, J.~Zhao, D.~Yu, N.~Du, I.~Shafran, K.~Narasimhan, and Y.~Cao, ``React: Synergizing reasoning and acting in language models,'' in \emph{International Conference on Learning Representations (ICLR)}, 2023.

\bibitem{han2024interpret}
M.~Han, Y.~Zhu, S.-C. Zhu, Y.~N. Wu, and Y.~Zhu, ``Interpret: Interactive predicate learning from language feedback for generalizable task planning,'' \emph{arXiv preprint arXiv:2405.19758}, 2024.

\bibitem{zawalski2024robotic}
M.~Zawalski, W.~Chen, K.~Pertsch, O.~Mees, C.~Finn, and S.~Levine, ``Robotic control via embodied chain-of-thought reasoning,'' \emph{arXiv preprint arXiv:2407.08693}, 2024.

\bibitem{zhang2024learning}
K.~Zhang, Z.-H. Yin, W.~Ye, and Y.~Gao, ``Learning manipulation skills through robot chain-of-thought with sparse failure guidance,'' \emph{arXiv preprint arXiv:2405.13573}, 2024.

\bibitem{wang2025large}
J.~Wang, E.~Shi, H.~Hu, C.~Ma, Y.~Liu, X.~Wang, Y.~Yao, X.~Liu, B.~Ge, and S.~Zhang, ``Large language models for robotics: Opportunities, challenges, and perspectives,'' \emph{Journal of Automation and Intelligence}, vol.~4, no.~1, pp. 52--64, 2025.

\bibitem{james2020rlbench}
S.~James, Z.~Ma, D.~R. Arrojo, and A.~J. Davison, ``Rlbench: The robot learning benchmark \& learning environment,'' \emph{IEEE Robotics and Automation Letters}, vol.~5, no.~2, pp. 3019--3026, 2020.

\bibitem{jiang2022vima}
Y.~Jiang, A.~Gupta, Z.~Zhang, G.~Wang, Y.~Dou, Y.~Chen, L.~Fei-Fei, A.~Anandkumar, Y.~Zhu, and L.~Fan, ``Vima: General robot manipulation with multimodal prompts,'' \emph{arXiv preprint arXiv:2210.03094}, vol.~2, no.~3, p.~6, 2022.

\bibitem{brohan2022rt}
A.~Brohan, N.~Brown, J.~Carbajal, Y.~Chebotar, J.~Dabis, C.~Finn, K.~Gopalakrishnan, K.~Hausman, A.~Herzog, J.~Hsu \emph{et~al.}, ``Rt-1: Robotics transformer for real-world control at scale,'' \emph{arXiv preprint arXiv:2212.06817}, 2022.

\bibitem{liu2023libero}
B.~Liu, Y.~Zhu, C.~Gao, Y.~Feng, Q.~Liu, Y.~Zhu, and P.~Stone, ``Libero: Benchmarking knowledge transfer for lifelong robot learning,'' \emph{Advances in Neural Information Processing Systems}, vol.~36, pp. 44\,776--44\,791, 2023.

\bibitem{mees2022calvin}
O.~Mees, L.~Hermann, E.~Rosete-Beas, and W.~Burgard, ``Calvin: A benchmark for language-conditioned policy learning for long-horizon robot manipulation tasks,'' \emph{IEEE Robotics and Automation Letters}, vol.~7, no.~3, pp. 7327--7334, 2022.

\bibitem{guo2025deepseek}
D.~Guo, D.~Yang, H.~Zhang, J.~Song, R.~Zhang, R.~Xu, Q.~Zhu, S.~Ma, P.~Wang, X.~Bi \emph{et~al.}, ``Deepseek-r1: Incentivizing reasoning capability in llms via reinforcement learning,'' \emph{arXiv preprint arXiv:2501.12948}, 2025.

\bibitem{openai_gpt5_system_card_2025}
{OpenAI}, ``{The GPT-5 System Card},'' August 2025.

\bibitem{hurst2024gpt}
A.~Hurst, A.~Lerer, A.~P. Goucher, A.~Perelman, A.~Ramesh, A.~Clark, A.~Ostrow, A.~Welihinda, A.~Hayes, A.~Radford \emph{et~al.}, ``Gpt-4o system card,'' \emph{arXiv preprint arXiv:2410.21276}, 2024.

\bibitem{pan2024dynathink}
J.~Pan, Y.~Zhang, C.~Zhang, Z.~Liu, H.~Wang, and H.~Li, ``Dynathink: Fast or slow? a dynamic decision-making framework for large language models,'' \emph{arXiv preprint arXiv:2407.01009}, 2024.

\end{thebibliography}
}

\end{document}